\documentclass[preprints,article,accept,pdftex,moreauthors]{Definitions/mdpi} 
\usepackage{makecell} 

\usepackage[ruled,linesnumbered]{algorithm2e}
\usepackage{placeins}

\firstpage{1} 
\pubvolume{1}
\issuenum{1}
\articlenumber{0}
\pubyear{2026}
\copyrightyear{2026}
\datereceived{ } 
\daterevised{ } 
\dateaccepted{ } 
\datepublished{ } 

\Title{Frequency-Enhanced Dual-Subspace Networks for Few-Shot Fine-Grained Image Classification}

\Author{Meijia Wang $^{1}$, Guochao Wang $^{1}$, Haozhen Chu $^{1}$, Bin Yao $^{1}$, Weichuan Zhang $^{1,}$*, Yuan Wang $^{1}$ and Junpo Yang $^{1}$}

\AuthorNames{Meijia Wang, Guochao Wang, Haozhen Chu, Bin Yao, Weichuan Zhang, Yuan Wang and Junpo Yang}

\address[1]{%
	$^{1}$ \quad School of Electronic Information and Artificial Intelligence, Shaanxi University of Science and Technology, Xi'an 710000, China}

\corres{Correspondence: zwc2003@163.com; Tel.: +86-158-0926-0366 (W.Z.)}

\abstract{Few-shot fine-grained image classification aims to recognize subcategories with high visual similarity using only a limited number of annotated samples. Existing metric learning-based methods typically rely solely on spatial domain features. Confined to this single perspective, models inevitably suffer from inherent texture biases, entangling essential structural details with high-frequency background noise. Furthermore, lacking cross-view geometric constraints, single-view metrics tend to overfit this noise, resulting in structural instability under few-shot conditions. To address these issues, this paper proposes the Frequency-Enhanced Dual-Subspace Network (FEDSNet). Specifically, FEDSNet utilizes the Discrete Cosine Transform (DCT) and a low-pass filtering mechanism to explicitly isolate low-frequency global structural components from spatial features, thereby suppressing background interference. Truncated Singular Value Decomposition (SVD) is employed to construct independent, low-rank linear subspaces for both spatial texture and frequency structural features. An adaptive gating mechanism is designed to dynamically fuse the projection distances from these dual views. This strategy leverages the structural stability of the frequency subspace to prevent the spatial subspace from overfitting to background features. Extensive experiments on four benchmark datasets—CUB-200-2011, Stanford Cars, Stanford Dogs, and FGVC-Aircraft—demonstrate that FEDSNet exhibits excellent classification performance and robustness, achieving highly competitive results compared to existing metric learning algorithms. Complexity analysis further confirms that the proposed network achieves a favorable balance between high accuracy and computational efficiency, providing an effective new paradigm for few-shot fine-grained visual recognition.}


\begin{document}

\section{Introduction}

While deep learning has significantly advanced generic image classification \cite{ref1,ref36,ref68}, Fine-Grained Visual Categorization (FGVC) remains a formidable challenge \cite{ref2}. Unlike general classification, FGVC aims to distinguish highly confusable subcategories within a broad category, such as specific bird species, car models, or aircraft types \cite{ref26,ref27,ref28}. This task is inherently difficult due to the minimal inter-class visual variance and substantial intra-class variations caused by diverse postures, lighting conditions, and background noise. Consequently, traditional feature extraction paradigms often struggle to capture the discriminative and subtle localized details required for accurate classification.

Furthermore, although modern FGVC algorithms achieve remarkable performance, they heavily rely on massive datasets annotated by domain experts. In practical scenarios, acquiring such high-quality, specialized annotations is prohibitively expensive and time-consuming. This critical bottleneck caused by data scarcity limits model generalization and inherently motivates the paradigm of Few-Shot Fine-Grained Image Classification (FS-FGIC). Mimicking the human cognitive ability to generalize from minimal observations, FS-FGIC aims to accurately recognize novel subcategories (the query set) using only a handful of labeled samples (the support set) \cite{ref3,ref4}. Currently, mainstream approaches in this domain are predominantly based on metric learning, which tackles the classification task by embedding images into a discriminative low-dimensional feature space and computing the semantic distance between the query and support samples \cite{ref3,ref4,ref5,ref6,ref69}.

However, existing metric learning-based FS-FGIC methods still face significant challenges in practical applications, primarily stemming from feature entanglement and the structural instability of their metric mechanisms. Specifically, this single-view paradigm often lacks sufficient constraints to fully decouple essential object structures from environmental noise. Specifically, this manifests as Feature Entanglement and Structural Instability:
\begin{enumerate}
\item At the feature representation level, current mainstream methods typically employ standard Convolutional Neural Networks (CNNs, e.g., ResNet-12) to map images into high-dimensional spatial features \cite{ref3,ref4,ref5,ref55,ref56,ref57}. Due to the inherent texture bias of CNNs \cite{ref48}, the structural details of fine-grained objects are frequently entangled with complex background clutter (e.g., grass or bushes). Consequently, the extracted feature tensors contain a substantial amount of non-discriminative, high-frequency noise. Even recent local alignment-based approaches \cite{ref6,ref9} struggle to disentangle this environmental noise under extreme few-shot constraints, inevitably leading to feature distraction.
 
\item At the metric construction level, current mainstream methods predominantly follow a prototype-based metric learning framework, which abstracts each category as a single feature point in the embedding space. However, such point estimation is often insufficient to characterize the complex intra-class distributions of fine-grained samples under varying poses and viewpoints. To capture higher-order geometric information, structural metric paradigms such as DSN \cite{ref10}, MetaOptNet \cite{ref50}, and CovaMNet \cite{ref51} have been introduced. These methods typically assume that samples belonging to the same class can be adequately represented by a compact linear subspace or a unimodal distribution, thereby providing higher degrees of freedom for modeling intra-class variations. Nevertheless, due to their single-view modeling mechanisms, existing structural metric paradigms exhibit restricted expressive power under extreme few-shot constraints. Fine-grained intra-class samples frequently display highly non-linear pose variations coupled with complex environmental interference. A single-view metric lacks the necessary constraints to decouple the intrinsic structural variations of the object from high-frequency noise. During the training phase, this single subspace tends to overfit local salient textures or high-frequency interference, causing it to deviate from the essential global structure of the object, as illustrated in Figure 1(a). Consequently, when query samples undergo pose variations or background shifts, this unconstrained single-view metric exhibits pronounced structural instability, ultimately resulting in significant performance degradation.
\end{enumerate}

To address these issues, this paper proposes the Frequency-Enhanced Dual-Subspace Network (FEDSNet). To mitigate the feature entanglement problem, a Frequency Structural Branch is designed. Unlike traditional methods that rely solely on spatial features, this approach utilizes the Discrete Cosine Transform (DCT) and low-pass filtering to explicitly extract the low-frequency energy of the image. This low-frequency information effectively delineates the global shape and contours of the object, complementing the local texture details extracted by the spatial branch (as shown in Figure 1(b)). Consequently, it achieves a fundamental decoupling of intrinsic structural information from background noise at the feature extraction stage \cite{ref10}.Furthermore, to overcome the instability problem at the metric level, this paper introduces a Structure-Constrained Dual-Subspace Metric. While recent methods like DSN \cite{ref10} have demonstrated the efficacy of Truncated SVD in constructing low-rank manifolds and discarding non-salient noise \cite{ref7,ref14}, applying decomposition to a single spatial feature remains highly susceptible to directional deviation. Therefore, to transcend the limitations of a single view, Truncated SVD is independently applied to the decoupled spatial texture features and frequency structural features to establish complementary linear subspaces. Crucially, the frequency subspace, characterized by its inherent low-frequency purity, acts as a 'Structural Anchor'. This anchor effectively calibrates the spatial subspace and mitigates its tendency to overfit background noise. Finally, robust fine-grained classification is achieved by adaptively weighting and fusing the subspace projection distances from both views.

\begin{figure}[H]
\centering
\includegraphics[width=\textwidth]{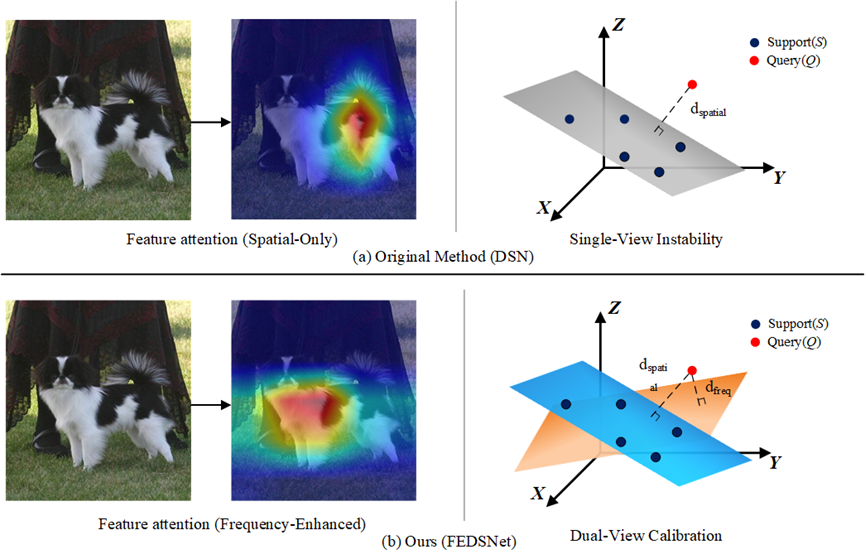} 

\caption{Comparison of feature representations and subspace manifolds. Visualizations of feature attention maps (left)(generated via Grad-CAM \cite{ref34}) and their corresponding geometric subspace structures (right) are presented. (a) Existing subspace methods (e.g., DSN) are highly susceptible to background interference. Due to feature entanglement, the constructed singular subspace is forced to overfit high-frequency background noise, failing to capture the intrinsic structure of the object. (b) Our method (FEDSNet) utilizes frequency constraints to explicitly extract global structural information. This effectively decouples the object from environmental clutter, concentrating the attention map strictly on the main body and generating calibrated, robust dual-subspaces for accurate classification.}

\label{fig:fig1}
\end{figure}

In summary, the main contributions of this paper are threefold:

\begin{enumerate}
    \item A novel dual-view feature decoupling mechanism: To address the vulnerability of CNN features to background texture interference, a parallel frequency structural enhancement branch is designed. By leveraging the Discrete Cosine Transform (DCT) and low-pass filtering, robust low-frequency structural components are explicitly isolated. This enable fundamental decoupling of spatial textures from frequency structures, thereby suppressing high-frequency noise at the feature extraction stage.
    
    \item A structure-constrained dual-subspace metric framework: Transcending the limitations of conventional single-subspace methods, independent discriminative subspaces are constructed for the decoupled spatial and frequency features, respectively. By adaptively fusing the projection distances from both views, the inherent structural stability of the frequency subspace is leveraged to explicitly calibrate the spatial subspace, achieving a highly robust modeling of intra-class manifolds.
    
\end{enumerate}
\section{Related Work}

\subsection{Few-Shot Learning}

Current research in Few-Shot Learning (FSL) \cite{ref18} has predominantly evolved along two distinct trajectories: optimization-based and metric-based paradigms.

Optimization-based approaches focus on meta-learning strategies to acquire task-agnostic initialization parameters for rapid adaptation. Representative works such as MAML \cite{ref19} and its variants \textit{e.g., Meta-SGD} \cite{ref20}, \textit{TAML} \cite{ref21} enable models to converge on novel tasks via a few gradient updates. Subsequent advancements like LEO \cite{ref37} and ANIL \cite{ref38} have sought to mitigate the high-dimensional optimization difficulties inherent in this process. However, despite their theoretical elegance, optimization-based methods frequently suffer from training instability, slow convergence, and prohibitive computational overhead --- particularly those necessitating second-order derivative calculations --- limiting their scalability in complex visual tasks.

Consequently, metric-based methods have emerged as the dominant paradigm due to their architectural simplicity and highly efficient inductive biases. The core philosophy of these methods involves constructing a discriminative embedding space where intra-class compactness and inter-class separability are maximized. Foundational works, including Siamese Networks \cite{ref22} and Triplet Networks \cite{ref23}, rely on pair-wise distance constraints. Subsequent milestones have established more sophisticated instance-to-class metric protocols, utilizing attention mechanisms (Matching Networks \cite{ref3}), class prototypes (Prototypical Networks \cite{ref4}), and learnable non-linear modules (Relation Networks \cite{ref5}). To further combat data scarcity, hallucination-based strategies \cite{ref39,ref40} attempt to augment feature distributions synthetically. Nevertheless, when applied to fine-grained recognition, these conventional metric and hallucination strategies often struggle to capture subtle discriminative details and are susceptible to artifact noise, thereby prompting the need for more structurally robust metric frameworks.

\subsection{Fine-Grained Few-Shot Classification}

Fine-Grained Visual Categorization (FGVC) necessitates the capture of subtle, localized discrepancies (e.g., beak shapes or headlight textures). Under few-shot constraints, holistic global features frequently fail to distinguish highly confusable subcategories. Consequently, existing literature predominantly explores localized feature matching, attention mechanisms, and recently, Vision Transformers (ViTs).

To capture intricate local details, a major line of research focuses on dense feature alignment and reconstruction. For local alignment, methods like DN4 \cite{ref6} introduce image-to-class local descriptor metrics, while DeepEMD \cite{ref9} leverages Earth Mover's Distance to formulate optimal transport for region matching. Similarly, ADM \cite{ref41} employs asymmetric distribution metrics. Alternatively, feature reconstruction approaches, such as FRN \cite{ref24} and Bi-FRN \cite{ref25}, utilize support features to linearly reconstruct queries, classifying based on reconstruction residuals. BSNet \cite{ref29} further refines this paradigm by capturing micro-discriminative traits via bi-similarity metrics.

Another prominent trajectory utilizes attention mechanisms to explicitly highlight discriminative regions. Inspired by standard attention modules \cite{ref42,ref43}, MattML \cite{ref15} and CTM \cite{ref44} design task-adaptive and category-specific embeddings, respectively. Dual Attention \cite{ref16} synergistically amplifies foreground targets across spatial and channel dimensions. However, recent analyses \cite{ref30,ref31, ref58, ref59, ref60, ref61, ref62, ref63, ref64,ref67} reveal a critical vulnerability: these spatial attention mechanisms remain highly susceptible to background noise and feature entanglement. Without explicit background suppression strategies, attention modules often erroneously focus on salient environmental clutter rather than the object itself.

Recently, Vision Transformers (ViTs) \cite{ref45} have been introduced to the FSL domain to model long-range dependencies (e.g., CrossTransformers \cite{ref32}, ViT-FSL \cite{ref33}, PMF \cite{ref46}). Despite their theoretical superiority in capturing global context, ViTs are inherently data-hungry. Under extreme few-shot conditions, they suffer from severe overfitting and impose prohibitive computational overhead. This reaffirms that effectively integrating the efficient inductive biases of CNNs with robust, explicit structural constraints remains the most viable and efficient paradigm for fine-grained problems.

\subsection{Subspace Learning and Low-Rank Constraints}

The manifold hypothesis posits that high-dimensional visual data inherently lie on low-dimensional sub-manifolds embedded within the ambient space. Unlike traditional metric methods (e.g., ProtoNet \cite{ref4}) that collapse entire categories into singular prototype points, subspace learning leverages the geometric structure of the support set to formulate more robust decision boundaries. As a representative framework, Deep Subspace Networks (DSN) \cite{ref10} postulates that samples from each class span an independent low-dimensional linear subspace. It employs truncated Singular Value Decomposition (SVD) to extract principal directions and classifies queries based on projection distances. Similarly, MetaOptNet \cite{ref50} applies convex optimization to construct a maximum-margin hyperplane, while CovaMNet \cite{ref51} utilizes second-order statistics (covariance matrices) to approximate the intra-class geometric distribution.

However, a critical limitation of these conventional subspace frameworks is their inherent hypersensitivity to outliers and non-salient features. In fine-grained scenarios, complex background clutter often manifests as high-variance directional components. Consequently, singular structures—whether the subspaces in DSN or the ellipsoids in CovaMNet—inevitably overfit these dominant background variations, yielding biased class representations. To mitigate this, low-rank constraint theories have been introduced to explicitly disentangle the clean intrinsic data structure from sparse noise. For example, SVDNet \cite{ref47} enforces feature decorrelation via orthogonal weight constraints, and Low-Rank Pairwise Alignment \cite{ref7} attempts to eliminate spatial redundancy during feature matching.

Nevertheless, a fundamental oversight in most existing low-rank methodologies is that they operate exclusively within the spatial domain. In complex scenes, spatial constraints often fail to decisively separate object structures from coherent background textures. Departing from these paradigms, our proposed method leverages the unique characteristics of the frequency domain. Acting as a spectral filter, it explicitly suppresses high-frequency background noise to guarantee representation purity at the source. This pivotal spectral decoupling ensures that the subsequently constructed spatial subspaces naturally and accurately approximate the essential, low-rank manifold structure of the categories.

\subsection{Learning in Frequency Domain}

While Convolutional Neural Networks (CNNs) have shown great effectiveness in feature extraction, standard architectures often exhibit a "texture bias" \cite{ref48}, tending to rely on local texture patches rather than global shape contours for classification. In fine-grained visual categorization tasks, this bias presents a specific challenge: although textures are discriminative, models frequently struggle to distinguish them from complex background textures (e.g., forests, grass), which may lead to non-semantic attention drift.

To capture global structural information more robustly, learning in the frequency domain has gained increasing attention. Xu et al.\cite{ref11} demonstrated that training directly in the discrete cosine transform (DCT) domain can maintain classification accuracy using low-frequency components while reducing computational costs. Similarly, GFNet \cite{ref12} utilizes the Fast Fourier Transform (FFT) to efficiently model long-range dependencies. Furthermore, FcaNet \cite{ref49} introduces frequency domain analysis into channel attention, indicating that different frequency components contain differentiated information: low-frequency components are often correlated with global structure, while high-frequency components encode local details.Moreover, frequency-aware clues have also demonstrated strong discriminative power in capturing subtle artifacts in complex visual tasks like face forgery detection \cite{ref13, ref65, ref66,ref70,ref71}.

Inspired by the aforementioned works, the dual-branch architecture proposed in this paper aims to separate structural information from texture details. Instead of discarding textures, we utilize the low-frequency component as a structural constraint to guide the model to focus on the global shape of the object. This mechanism helps to calibrate the spatial branch, enabling it to better distinguish meaningful object details from background interference. Therefore, this paper introduces a frequency-based structural prior into the metric construction for few-shot fine-grained classification.

\section{Method}
\subsection{Problem Definition}

Formally, let $\mathcal{D} = (x_i, y_i)$ denote a dataset, where $x_i$ represents an input image and $y_i \in \mathcal{C}$ represents its corresponding class label. The dataset $\mathcal{D}$ is partitioned into three mutually disjoint subsets: a base dataset $\mathcal{D}_{base}$, a validation dataset $\mathcal{D}_{val}$, and a novel dataset $\mathcal{D}_{novel}$. Their respective label spaces, $\mathcal{C}_{base}$, $\mathcal{C}_{val}$, and $\mathcal{C}_{novel}$, are strictly non-overlapping (\textit{i.e.}, $\mathcal{C}_{base} \cap \mathcal{C}_{val} = \mathcal{C}_{val} \cap \mathcal{C}_{novel} = \mathcal{C}_{base} \cap \mathcal{C}_{novel} = \emptyset$), ensuring that the classes encountered during testing are entirely unseen during training.

To simulate few-shot scenarios and promote model generalization, we adopt the standard episodic training paradigm. During the meta-training phase, tasks are iteratively sampled from $\mathcal{D}_{base}$, while in the meta-testing phase, the evaluation is performed by sampling from $\mathcal{D}_{novel}$. Each sampled task (or episode) is formulated as an $N$-way $K$-shot classification problem, which comprises a support set $\mathcal{S}$ and a query set $\mathcal{Q}$:

\noindent \textbf{Support Set $\mathcal{S}$:} Consists of $N$ classes randomly sampled from the current dataset, with $K$ labeled instances drawn per class, yielding $\mathcal{S}=\{(x_s, y_s)\}_{s=1}^{N \times K}$.

\noindent \textbf{Query Set $\mathcal{Q}$:} Comprises $M$ instances randomly drawn from the same $N$ classes to evaluate task-specific performance, yielding $\mathcal{Q}=\{(x_q, y_q)\}_{q=1}^{N \times M}$.

The fundamental objective of this episodic formulation is to leverage the abundant data in $\mathcal{D}_{base}$ to learn a transferable metric space. This enables the model to accurately predict the labels of query samples in $\mathcal{D}_{novel}$ by rapidly adapting to the minimal labeled examples provided in the support set $\mathcal{S}$.

\subsection{Overall Framework}

As illustrated in Figure 2, this paper proposes the Frequency-Enhanced Dual-Subspace Network (FEDSNet). By integrating frequency-domain decoupling with low-rank subspace learning, this framework aims to systematically address the persistent challenges of feature entanglement, underutilization of discriminative information, and few-shot overfitting in fine-grained categorization.

\begin{figure}[H]
\centering
\includegraphics[width=\textwidth]{figure2.png} 

\caption{The overall architecture of the proposed FEDSNet framework. Input images are first encoded by a shared backbone (e.g., ResNet-12) to extract spatial feature maps. Subsequently, the pipeline diverges into two complementary branches: (1) Frequency-aware Decomposition Module: In the frequency stream, robust global structural features are explicitly decoupled from high-frequency noise via the Discrete Cosine Transform (DCT), low-pass filtering, and frequency attention mechanisms. (2) Low-Rank Multi-Subspace Module: For both spatial and frequency streams, truncated Singular Value Decomposition (SVD) is employed to construct class-specific, low-rank subspaces. Finally, the projection distances of the query image to the dual subspaces are fused via adaptive weights $\alpha$ to yield the final prediction.}

\label{fig:fig2}
\end{figure}

\noindent For a given $N$-way $K$-shot task, the procedural workflow is formulated through the following three primary stages:

\begin{enumerate}[label=\arabic*., leftmargin=*, labelsep=2mm]
    \item \textbf{Feature Extraction:} Images from both the support set $\mathcal{S}$ and query set $\mathcal{Q}$ are first processed by a shared backbone network (e.g., ResNet-12 or Conv-4). This stage yields raw spatial feature tensors that capture initial visual representations.
    
    \item \textbf{Frequency-aware Feature Decoupling:} To disentangle discriminative fine-grained details from common background clutter, the raw features are fed into the Frequency-aware Decomposition Module. Specifically, the Discrete Cosine Transform (DCT) is utilized to project the features into the frequency domain. A low-pass filtering mechanism is then applied to isolate robust low-frequency structural components. Subsequently, a frequency attention mechanism dynamically recalibrates channel-wise importance to emphasize critical information. Finally, the optimized spectrum is mapped back to the spatial domain via the Inverse DCT (IDCT), generating denoised structural features.
    
    \item \textbf{Low-Rank Multi-Subspace Metric:} To leverage the complementary information, independent Low-Rank Multi-Subspace Modules are constructed for both the raw spatial and the structure-enhanced feature views. Diverging from traditional point-prototype approaches, this module establishes a dedicated linear subspace for each class within each feature view. Truncated Singular Value Decomposition (SVD) is introduced as a low-rank constraint to effectively suppress redundant noise and recover the underlying manifold structure. This results in a comprehensive set of subspaces that encompass both localized texture details and holistic global structures. During inference, query samples are classified according to their projection distances to these subspaces. By employing an adaptive weighted fusion mechanism to integrate distance metrics from the dual views, the model generates the final category probabilities. The entire framework is optimized end-to-end via a joint objective function comprising classification loss and low-rank regularization.
\end{enumerate}

\subsection{Frequency-aware Decomposition Module}

While raw spatial features extracted by backbone networks contain rich texture details, they are often entangled with background clutter and are highly sensitive to pose variations in few-shot scenarios. To extract robust global geometric structures from limited samples, a frequency structural enhancement branch is designed. This branch explicitly segregates stable morphological information from noisy features through explicit spectral filtering and an attention mechanism. The specific process involves the following three key steps:

Spectrum Transformation and Low-Pass Filtering Given an input feature tensor $X \in \mathbb{R}^{C \times H \times W}$, a 2D Discrete Cosine Transform (2D-DCT) is first applied independently across each channel to map it from the spatial domain to the frequency domain. For a feature map with spatial coordinates $(x, y)$, its transformation to frequency coordinates $(u, v)$ is defined as:

\begin{linenomath}
\begin{equation}
F_c(u, v) = \alpha(u)\alpha(v) \sum_{x=0}^{H-1} \sum_{y=0}^{W-1} X_c(x, y) \cos \left[ \frac{\pi(2x + 1)u}{2H} \right] \cos \left[ \frac{\pi(2y + 1)v}{2W} \right],
\end{equation}
\end{linenomath}

where $F_c(u, v)$ represents the frequency spectrum, and the normalization compensation coefficients $\alpha(u)$ and $\alpha(v)$ are defined such that when $u, v = 0$, the coefficient is $1/\sqrt{H}$ or $1/\sqrt{W}$, and otherwise $\sqrt{2/H}$ or $\sqrt{2/W}$. The introduction of this transformation aims to achieve explicit decoupling of spatial features. In frequency-domain signal processing, although the Fast Fourier Transform (FFT) is a classic analytical tool, this study prioritizes the 2D-DCT primarily due to two irreplaceable advantages: First, its compatibility with deep architectures through purely real-number operations. The output of FFT inevitably introduces the complex domain; forcibly stripping the phase spectrum would severely disrupt the global spatial topological structure of fine-grained targets, while retaining the phase is difficult to directly integrate with existing real-domain network layers. Conversely, DCT relies solely on real cosine bases and can be perfectly embedded into end-to-end frameworks with minimal architectural overhead. Second, its energy compaction property. For highly correlated deep features of natural images, DCT can more efficiently compress and aggregate the low-frequency structural energy---representing smooth contours and holistic shapes of objects---into the top-left corner of the spectrum (i.e., the low-frequency index region where $u$ and $v$ are small), while pushing high-frequency redundant noise, which is susceptible to background interference, toward the bottom-right corner.

Leveraging this physical position decoupling property, a low-pass mask $M_{\mathrm{low}}$ based on normalized Manhattan distance is designed to precisely truncate high-frequency background noise. For any frequency coordinate $(u, v)$, the indicator function of this mask is defined as:

\begin{linenomath}
\begin{equation}
M_{\mathrm{low}}(u, v) = 
\begin{cases} 
1, & \text{if } \frac{1}{2}\left(\frac{u}{H} + \frac{v}{W}\right) \le \tau, \\ 
0, & \text{otherwise},
\end{cases}
\end{equation}
\end{linenomath}
where $\tau \in (0,1)$ is a predefined cutoff threshold (default set to 0.3). When the coordinate falls within the low-frequency triangular region formed by the cutoff threshold, the mask value is 1 (retaining the frequency band); otherwise, it is 0 (filtering the frequency band). By element-wise multiplying this mask with the original frequency spectrum, a preliminarily pure low-frequency structural feature is obtained:

\begin{linenomath}
\begin{equation}
F_{\mathrm{low}} = F \odot M_{\mathrm{low}},
\end{equation}
\end{linenomath}
where $\odot$ denotes element-wise multiplication.

After preserving the low-frequency structural information, considering that different frequency channels contain differentiated information regarding the object's structure, this module introduces a frequency channel attention mechanism to adaptively reweight the feature channels. Traditional channel attention typically utilizes fully connected layers and ReLU activation directly; however, in the frequency feature space, the energy distribution across different frequency channels exhibits massive magnitude disparities, forming a long-tail distribution. Direct linear mapping under this condition can easily cause the network to suffer from gradient explosion or vanishing during backpropagation. Therefore, after aggregating spatial energy via Global Average Pooling (GAP), this study introduces $\mathrm{LayerNorm}$ for energy scale normalization, and employs a Leaky ReLU with a negative slope of 0.1 to prevent neuron death in low-energy channels. The generated channel weight vector $w_{\mathrm{attn}} \in \mathbb{R}^C$ is formulated as:

\begin{linenomath}
\begin{equation}
w_{\mathrm{attn}} = \sigma \left( W_2 \delta_{LReLU} (W_1 \mathrm{LayerNorm}(\mathrm{GAP}(F_{\mathrm{low}}))) \right),
\end{equation}
\end{linenomath}

where $\sigma$ denotes the Sigmoid function, and $\delta_{\mathrm{LReLU}}$ is the Leaky ReLU function. The parameters $W_1 \in \mathbb{R}^{\frac{C}{r} \times C}$ and $W_2 \in \mathbb{R}^{C \times \frac{C}{r}}$ represent the weight matrices for the dimensionality reduction and expansion layers, respectively, with $r$ denoting the reduction ratio. Applying this weight vector to the frequency features yields the enhanced spectrum:

\begin{linenomath}
\begin{equation}
F'_{\mathrm{low}} = F_{\mathrm{low}} \odot w_{\mathrm{attn}}.
\end{equation}
\end{linenomath}

This mechanism enables the model to stably and adaptively filter the feature channels that are most critical for representing the essential structure of the object. 

Finally, to align the enhanced frequency information with the original spatial features to construct unified-dimension subspaces later, a 2D Inverse Discrete Cosine Transform (2D-IDCT) is utilized to restore $F'_{\mathrm{low}}$ back to the spatial domain, yielding the final structure-enhanced features: $X_{\mathrm{shape}} = \mathrm{IDCT}(F'_{\mathrm{low}})$. Compared to the raw features $X$, $X_{\mathrm{shape}}$ successfully filters out redundant high-frequency textures and background clutter, focusing intensely on the essential morphology and topological consistency of the object, thereby providing an extremely robust structural constraint for subsequent classification. The complete workflow of this module is summarized in Algorithm~\ref{algorithm}

\begin{algorithm}[htb] 
	\caption{Frequency-aware Decomposition Module}
	\label{algorithm}
	\KwData{Raw spatial feature tensor \(X \in \mathbb{R}^{C \times H \times W}\); Cutoff threshold \(\tau \in (0, 1)\); Learnable parameter matrices \(W_1\) and \(W_2\).}
	\KwResult{Structure-enhanced feature tensor \(X_{shape} \in \mathbb{R}^{C \times H \times W}\)}
	
	\textbf{Step 1: Spectrum Transformation:}
	
	Apply 2D Discrete Cosine Transform (2D-DCT) to each channel of \(X\) to obtain the frequency spectrum \(F\):
	
	\(F \gets \text{2D-DCT}(X)\)
	
	\textbf{Step 2: Low-pass Filtering:}
	
	Initialize the low-pass mask \(M_{low}\) as a tensor of size \(C \times H \times W\) filled with \(0\).
	
	\For{\(u = 0\) to \(H-1\)} {
		\For{\(v = 0\) to \(W-1\)} {
			\eIf{\(\frac{1}{2}\left(\frac{u}{H} + \frac{v}{W}\right) \leq \tau\)}{
				\(M_{low}[:, u, v] \gets 1\)
			}{
				\(M_{low}[:, u, v] \gets 0\)
			}
		}
	}
	
	Extract the low-frequency structural components via element-wise multiplication (\(\odot\)):
	
	\(F_{low} \gets F \odot M_{low}\)
	
	\textbf{Step 3: Frequency Channel Attention:}
	
	Aggregate spatial energy via Global Average Pooling (GAP):
	
	\(E \gets \text{GAP}(F_{low})\)
	
	Compute channel attention weights \(w_{attn}\) using the bottleneck structure:
	
	\(w_{attn} \gets \sigma\left( W_{2} \delta_{LReLU}\left( W_{1} \text{LayerNorm}(E) \right) \right)\)
	
	Re-weight the low-frequency features:
	
	\(F'_{low} \gets F_{low} \odot w_{attn}\)
	
	\textbf{Step 4: Spatial Domain Restoration:}
	
	Apply 2D Inverse Discrete Cosine Transform (2D-IDCT) to restore the enhanced spectrum to the spatial domain:
	
	\(X_{shape} \gets \text{2D-IDCT}(F'_{low})\)
	
	\textbf{Step 5: Return Result:}
	
	\Return \(X_{shape}\)
\end{algorithm}

\subsection{Low-Rank Multi-Subspace Construction and Adaptive Metric}

To address the inherent structural instability of single spatial subspaces when facing complex background interference, this paper proposes a structure-constrained dual-subspace metric mechanism. Specifically, instead of relying on a single view, this mechanism constructs independent, low-rank class subspaces for both the spatial domain (retaining rich local textures) and the frequency domain (providing denoised global structures), respectively. By adaptively fusing the projection distances from these dual domains, it explicitly leverages the frequency structure to calibrate the spatial metric, dynamically balancing their contributions to achieve robust few-shot classification.

Traditional metric learning methods (e.g., Prototypical Networks) typically utilize feature means to represent categories. However, this point estimation neglects the structural distribution of intra-class data. Inspired by the DSN framework, this study adopts a subspace modeling paradigm, treating each category as a low-dimensional linear manifold. Compared to point prototypes, subspaces are capable of capturing the principal directions of intra-class variations under varying poses and viewpoints, thereby providing higher-order discriminative information.

\subsubsection{Subspace Construction via Truncated SVD}

For a given category $k$ and feature view $v \in \{\text{spatial, shape}\}$, let its support set feature matrix be $S_{k,v} \in \mathbb{R}^{D \times N}$, where $D$ is the feature dimension and $N$ is the number of samples. Initially, a centering operation is performed by subtracting the class mean $\mu_{k,v}$ from each sample, yielding the centered matrix:

\begin{linenomath}
\begin{equation}
\bar{S}_{k,v} = S_{k,v} - \mu_{k,v}.
\end{equation}
\end{linenomath}

This step eliminates coordinate offset interference in the high-dimensional space, ensuring that the singular values extracted subsequently accurately reflect the data variance along each basis direction. 

While SVD is effective for manifold construction, under extreme few-shot settings (e.g., 1-shot) or when data augmentation is applied, the feature matrix is prone to rank deficiency or high collinearity. This can lead to numerical instability during decomposition. To address this, we introduce micro-Gaussian noise (scaled by $10^{-5}$) to the support set matrix as a numerical stabilization mechanism to prevent rank deficiency and ensure the convergence of SVD:

\begin{linenomath}
\begin{equation}
\tilde{S}_{k,v} = \bar{S}_{k,v} + \epsilon_{\text{jitter}}, \quad \epsilon_{\text{jitter}} \sim \mathcal{N}(0, 10^{-5}I).
\end{equation}
\end{linenomath}

\begin{figure}[H]
\centering
\includegraphics[width=\textwidth]{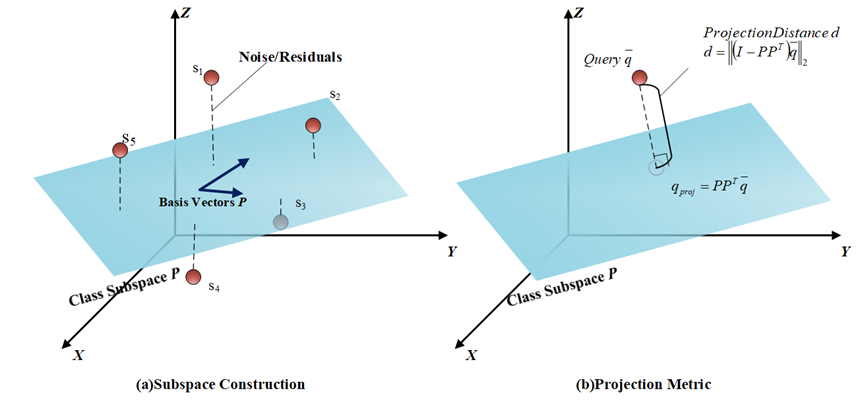} 
\caption{Geometric interpretation of subspace creation and projection metric. (\textbf{a}) Subspace Construction: Given a 5-shot support set (green spheres) containing instance-specific noise, Truncated SVD constructs a low-rank class subspace $P$ (blue plane). This subspace captures the principal data manifold spanned by the basis vectors $U$, filtering out-of-plane noise. (\textbf{b}) Projection Metric: To classify a query sample $q$ (red sphere), it is orthogonally projected onto the subspace to obtain its denoised reconstruction $q_{\mathrm{proj}}$. Similarity is measured by the projection distance $d \rightarrow$ (i.e., the reconstruction error $\|(I - UU^{\mathrm{T}})q\|_2$). A smaller distance indicates a better structural match with the class.}
\label{fig3}
\end{figure}

This operation not only structurally prevents division-by-zero errors but also serves as an implicit Tikhonov regularization for the low-rank manifold, enhancing robustness against perturbations. Subsequently, Singular Value Decomposition is applied to the noise-injected matrix:

\begin{linenomath}
\begin{equation}
\tilde{S}_{k,v} = U_{k,v} \Sigma_{k,v} V_{k,v}^{\mathrm{T}}.
\end{equation}
\end{linenomath}

where $U_{k,v}$ represents the left singular vector matrix, corresponding to the principal directions of the feature space. In fine-grained tasks, the tail singular values typically correspond to background noise or non-salient variations \cite{ref35}. Following the truncated SVD strategy, we retain only the top $d$ left singular vectors to construct the truncated orthogonal basis matrix $P_{k,v} = U_{k,v}[:, 1:d]$, as illustrated in Figure~\ref{fig3}. Unlike previous methods that perform SVD solely on spatial features, we apply this low-rank truncation independently to both feature views. Consequently, each category is modeled by two low-rank linear manifolds: the spatial texture subspace $P_{k,\mathrm{spatial}}$ and the structural shape subspace $P_{k,\mathrm{shape}}$. This approximation compels the model to disregard subtle noise and evaluate metrics exclusively on the principal component manifold representing the essential data structure.

\subsubsection{Adaptive Projection Metric}

During inference, given a query sample $q$, classification relies on its geometric projection distance to each category's subspace. To align the query sample and the category subspace within the same relative coordinate system, a class-specific centered query vector $\bar{q}_{k,v} = q - \mu_{k,v}$ is computed. According to the orthogonal projection theorem, the matrix $P_{k,v}P_{k,v}^{\mathrm{T}}$ constitutes the projection operator onto the subspace. The projection distance $d$ is formulated as the squared 2-norm of the residual in the orthogonal complement space:

\begin{linenomath}
\begin{equation}
d(q, P_{k,v}) = \left\| (I - P_{k,v}P_{k,v}^{\mathrm{T}})(q - \mu_{k,v}) \right\|_2^2.
\end{equation}
\end{linenomath}

For backpropagation, this distance must be converted into a similarity score $s$. In extreme cases where few-shot features highly overlap, $d \rightarrow 0$ can cause gradient explosion during square root derivation. Thus, a minuscule smoothing factor $\epsilon = 10^{-6}$ is introduced:

\begin{linenomath}
\begin{equation}
s(q, P_{k,v}) = -\sqrt{d(q, P_{k,v}) + \epsilon}.
\end{equation}
\end{linenomath}

The parameter $\epsilon$ applies a smooth truncation to the floor of the reconstruction error, allowing gradients to decay smoothly when a sample perfectly fits the subspace, thereby improving optimization stability. 

Considering that distinct fine-grained categories exhibit varying dependencies on texture details versus overall structure, a simple average fusion is sub-optimal. We introduce learnable fusion parameters $w = [w_{\mathrm{spatial}}, w_{\mathrm{shape}}]$, normalized via the $\mathrm{Softmax}$ function, to dynamically adjust the contribution weights $\alpha_v$:

\begin{linenomath}
\begin{equation}
\alpha_v = \frac{e^{w_v}}{\sum_{j \in \{\text{spatial, shape}\}} e^{w_j}}.
\end{equation}
\end{linenomath}

The final fused distance metric $\mathcal{D}(q, k)$ is the weighted sum of the similarities from both views:

\begin{linenomath}
\begin{equation}
\mathcal{D}(q, k) = \alpha_{\mathrm{spatial}} \cdot d(q, P_{k,\mathrm{spatial}}) + \alpha_{\mathrm{shape}} \cdot d(q, P_{k,\mathrm{shape}}).
\end{equation}
\end{linenomath}

This adaptive mechanism enables the model to dynamically find an optimal balance between capturing local textures and extracting global contours based on specific task requirements.

\subsubsection{Optimization Objective}

To further enhance subspace discriminability, we introduce a discriminative regularization term $L_{\mathrm{disc}}$ alongside the standard classification cross-entropy loss $L_{\mathrm{cls}}$. This term aims to maximize orthogonality between different category subspaces and minimize inter-class overlap:

\begin{linenomath}
\begin{equation}
L_{\mathrm{disc}} = \sum_{i \neq j} \| P_i^{\mathrm{T}}P_j \|_{\mathrm{F}}^2,
\end{equation}
\end{linenomath}
where $\| \cdot \|_{\mathrm{F}}$ denotes the Frobenius norm. By penalizing the inner product of basis vectors from distinct categories, $L_{\mathrm{disc}}$ explicitly forces the manifolds to be maximally orthogonal, enlarging decision margins between easily confusable categories. The total optimization objective is defined as:

\begin{linenomath}
\begin{equation}
L_{\mathrm{total}} = L_{\mathrm{cls}} + \lambda L_{\mathrm{disc}},
\end{equation}
\end{linenomath}
where $\lambda$ is a balancing coefficient. The entire framework is trained end-to-end, jointly optimizing the backbone parameters, attention weights, and metric fusion coefficients.

\section{Experiment}

\subsection{Datasets and Implementation Details}

To verify the effectiveness of the proposed method, experiments are evaluated on four widely used fine-grained few-shot benchmark datasets: CUB-200-2011 (birds), Stanford Dogs, Stanford Cars, and FGVC-Aircraft. This study strictly follows the standard data split protocols in the fine-grained few-shot domain. Specifically, the 200 classes of CUB-200-2011 are divided into 100, 50, and 50 classes for training, validation, and testing, respectively; the split ratios for Stanford Dogs, Stanford Cars, and FGVC-Aircraft are 60/30/30, 98/49/49, and 50/25/25, respectively. All input images are uniformly resized to a resolution of $84 \times 84$.

Regarding feature extraction, to ensure fair comparisons and verify the generality of the method, this paper adopts Conv-4 and ResNet-12 as backbone networks, respectively. Conv-4 consists of four convolutional blocks, each comprising 64 $3 \times 3$ convolution kernels, a $\mathrm{BatchNorm}$ layer, a $\mathrm{ReLU}$ activation function, and a $2 \times 2$ max-pooling layer. ResNet-12 contains four residual blocks with channel numbers of 64, 160, 320, and 640, respectively. To address the numerical distribution differences in the output features of different backbone networks and to prevent gradient saturation or vanishing during $\mathrm{Softmax}$ computation, the model sets learnable scale factors initialized at 1.0 and 10.0 for Conv-4 and ResNet-12, respectively.

Regarding the optimizer and hyperparameter settings, this paper adopts differentiated training strategies for different backbone networks. For the Conv-4 backbone, the model uses the Adam optimizer with an initial learning rate set to 0.001; for the ResNet-12 backbone, the model employs the SGD optimizer with Nesterov momentum, and the initial learning rate is set to 0.05. The weight decay for both is uniformly set to $5 \times 10^{-4}$. The meta-training phase is conducted for a total of 400 epochs, and the learning rate is decayed in 3 stages throughout the training process, with a decay coefficient ($\gamma$) of 0.1 each time. The model is evaluated on the validation set every 20~epochs. To ensure a smooth transition in the early stages of training, the learnable parameters for the dual-view metric fusion are uniformly initialized to 1.0, initially providing a balanced 1:1 contribution ratio before the model smoothly adapts to dynamic adaptive fusion. In subspace construction, the retained dimensions of truncated SVD are limited by the number of samples; for 1-shot and 5-shot tasks, the principal component dimensions are truncated to a maximum of 1 and 5, respectively. The balancing coefficient $\lambda$ for the discriminative orthogonal loss in joint optimization is strictly set to 0.03.

Additionally, for numerical stabilization during metric computation (as formulated in Section~3), the support set jitter scale and the distance smoothing factor $\epsilon$ are set to $10^{-5}$ and $10^{-6}$, respectively. The specific magnitudes of these parameters ($\lambda$, jitter scale, and $\epsilon$) were finalized through preliminary grid searches on the validation set. These exact values were empirically verified to be optimal sweet spots: smaller stabilization scales approach the precision limits of standard 32-bit floating-point operations (FP32), failing to prevent zero-division or SVD non-convergence, while larger perturbations or excessive orthogonal constraints risk masking the subtle intra-class variances critical for fine-grained feature learning.

The experimental settings cover standard $5$-way $1$-shot and $5$-way $5$-shot classification tasks. In each episode, in addition to the support set, $15$ query images are additionally sampled per class for evaluation. For the frequency branch, the cutoff threshold of the low-pass filter is set to $0.3$ by default. During the testing phase, the final performance is reported as the average classification accuracy over $600$ randomly sampled episodes, accompanied by a $95\%$ confidence interval. The code is implemented based on the PyTorch framework and trained and tested on an NVIDIA RTX $3090$ GPU.

\subsection{Performance Comparison}

To comprehensively evaluate the effectiveness of the proposed method (FEDSNet), this section conducts extensive comparisons with current mainstream few-shot metric learning baselines and state-of-the-art models under standard $5$-way $1$-shot and $5$-way $5$-shot settings. The compared models encompass classical point metric methods (e.g., ProtoNet ), local feature alignment-based models (e.g., DN4 , DeepEMD ), models focusing on spatial structural alignment (e.g., OLSA ), and recent feature reconstruction networks (e.g., FRN ). The baseline model is the original Deep Subspace Network (DSN) . The experiments are conducted on both shallow (Conv-4) and deep (ResNet-12) backbone networks. Detailed comparison results are summarized in Table~\ref{tab1}.

\begin{table}[H]
	\caption{Experimental comparison results of various methods on CUB-200-2011, FGVC-Aircraft, Stanford Dogs, and Stanford Cars datasets under different backbone networks. The best performance is highlighted in bold.\label{tab1}}
    \setlength{\tabcolsep}{1.2pt} 
    \makebox[\textwidth][c]{  
		\newcolumntype{C}{>{\centering\arraybackslash}X}
		\newcolumntype{M}{>{\centering\arraybackslash}p{\dimexpr0.2\textwidth-2\tabcolsep-\arrayrulewidth\relax}}
		\begin{tabularx}{\fulllength}{CMCCCCCCCC}
			\toprule
			\multirow{2}{*}{\textbf{Backbone}}	& \multirow{2}{*}{\textbf{Methods}}	& \multicolumn{2}{c}{\textbf{CUB-200-2011}}     & \multicolumn{2}{c}{\textbf{Aircraft}}     & \multicolumn{2}{c}{\textbf{Stanford Dogs}}	&\multicolumn{2}{c}{\textbf{Stanford Cars}}\\
			\cline{3-10}
			&  & \textbf{1-shot} & \textbf{5-shot} & \textbf{1-shot} & \textbf{5-shot}  & \textbf{1-shot} & \textbf{5-shot}  & \textbf{1-shot} & \textbf{5-shot}\\ 
			\midrule
			\multirow[m]{7}{*}{Conv-4}	& ProtoNet~\cite{ref4}			& 61.76±0.23			& 83.07±0.15			& -			& -			& 46.66±0.22			& 70.93±0.16			& 50.57±0.22			& 74.44±0.17\\
			& DN4~\cite{ref6}       & 57.45±0.89      & 84.41±0.58     & -   & -   & 39.08±0.76   & 69.81±0.69   & 34.12±0.68			& \textbf{87.47±0.47}\\
			& DeepEMD~\cite{ref9}       & 64.08±0.50      & 80.55±0.71     & -   & -   & 46.73±0.49   & 65.74±0.63   & 61.63±0.27			& 72.95±0.38\\
			& MattML~\cite{ref15}       & 66.29±0.56      & 80.34±0.30     & -   & -   & 54.84±0.53   & 71.34±0.38   & \textbf{66.11±0.54}			& 82.80±0.28   \\
			& MixFSL~\cite{ref8}       & 53.61±0.88      & 73.24±0.75     & 44.89±0.75   & 62.81±0.73   & -   & -   & 44.56±0.80			& 59.63±0.79\\
			& DSN~\cite{ref10}			& 65.86±0.23      & 83.80±0.15     & 49.70±0.20   & 65.67±0.18   & 53.98±0.22   & 64.62±0.18			& 40.17±0.21			&65.19±0.75\\
			& \textbf{FEDSNet (Ours)}			& \textbf{67.70±0.22}			& \textbf{84.92±0.15}			& \textbf{52.13±0.22}			& \textbf{69.37±0.18}			& \textbf{55.97±0.22}			& \textbf{75.18±0.16}			& 60.42±0.21			& 80.20±0.15\\
			\midrule
			\multirow[m]{9}{*}{ResNet-12} 
			& DeepEMD~\cite{ref9}			& 75.59±0.30			& 88.23±0.18			&  - 			& -			& 70.38±0.30			& 85.24±0.18			& 80.62±0.26			& 92.63±0.13\\  
			& LMPNet~\cite{ref54}			& -			& -			& -			& -			& 61.89±0.10			& 68.21±0.11			& 68.31±0.45			& 80.27±0.23\\
			& MixFSL~\cite{ref8}    & 67.87±0.94       & 82.18±0.66     & 60.55±0.86   & 77.57±0.69   & -   & -   	& 58.15±0.87			& 80.54±0.63\\
			& OLSA~\cite{ref17}   &  77.77±0.44     & 89.87±0.24     & -   &  -   & 64.15±0.49   & 78.28±0.32   	& 77.03±0.46			& 88.85±0.46\\
			& HelixFormer~\cite{ref52}		 & 81.66±0.30     & \textbf{91.83±0.17}     & -   & -   & 65.92±0.49   & 80.65±0.36			& 79.40±0.43			& 92.26±0.15\\
			& TOAN~\cite{ref53}   &  66.10±0.86     & 82.27±0.60     & -   &  -   & 49.77±0.86   & 69.29±0.70   	& 75.28±0.72			& 87.45±0.48\\
			& BSFA~\cite{ref31}   & \textbf{82.27±0.46}      & 90.76±0.26     & -   & -   & 69.58±0.50   & 82.59±0.33			& \textbf{88.93±0.38}			& 95.20±0.20\\
			& DSN (Base)~\cite{ref10}		 & 75.48±0.22      & 90.55±0.12     & 65.17±0.23   & 86.15±0.13   & 69.76±0.22   & 81.51±0.15			& 83.96±0.19			& 94.03±0.09\\
			& \textbf{FEDSNet (Ours)}			& 80.23±0.20			& 90.78±0.11			& \textbf{65.72±0.23}			& \textbf{86.33±0.13}			& \textbf{70.85±0.22}			& \textbf{86.87±0.13}			& 85.04±0.19			& \textbf{95.75±0.07}\\
			\bottomrule
		\end{tabularx}
    } 
\end{table}

As shown in Table~\ref{tab1}, the proposed FEDSNet method is comprehensively compared with a series of classical and recent few-shot learning models. The experimental results demonstrate that FEDSNet exhibits highly competitive classification performance and robustness across different network depths and fine-grained tasks. Notably, FEDSNet consistently outperforms the baseline DSN model across all datasets and network backbones (Conv-4 and ResNet-12). This performance improvement indicates that the proposed dual-subspace metric mechanism, which incorporates frequency feature decoupling and structural constraints, effectively mitigates the structural deficiencies of single-spatial-domain subspaces that are susceptible to background noise interference.

Under the Conv-4 backbone, which possesses a restricted parameter capacity, FEDSNet demonstrates a clear advantage in low-resource feature extraction. Particularly on the FGVC-Aircraft and Stanford Dogs datasets, FEDSNet achieves accuracies of $50.83\%$ and $55.97\%$ in 1-shot tasks, respectively, outperforming not only classical methods like ProtoNet and DN4 but also models relying on complex attention mechanisms, such as MattML. This advantage is primarily attributed to the introduction of the frequency structural enhancement branch at the feature extraction stage. When the shallow receptive field is insufficient to fully capture high-dimensional semantics, the global contour information explicitly retained by low-pass filtering provides the model with strong morphological priors, thereby avoiding overfitting to local high-frequency noise under extreme few-shot (1-shot) conditions.

When utilizing the deeper ResNet-12 backbone, alongside the enhanced spatial feature representation capabilities, the dual-view fusion mechanism of FEDSNet further improves performance. In the 5-shot task on Stanford Cars, FEDSNet achieves an accuracy of $95.75\%$, surpassing recent competitive models with complex structures, such as BSFA. Furthermore, on CUB-200-2011 and FGVC-Aircraft, FEDSNet achieves results that are highly competitive with, or superior to, the Vision Transformer-based HelixFormer. Compared to the complex bilateral semantic fusion of BSFA or the substantial self-attention computational overhead of HelixFormer, FEDSNet maintains a concise linear subspace metric architecture. Through truncated singular value decomposition and frequency domain calibration, FEDSNet provides a precise approximation of the intrinsic low-rank data manifold at a minimal computational cost, striking an effective balance between performance and efficiency.

Although FEDSNet yields optimal or highly competitive results in most scenarios, its performance on the Stanford Cars task under the Conv-4 backbone is slightly inferior to models specifically optimized for local salient features (e.g., MattML). This is likely because the fine-grained discrimination of car categories heavily relies on extremely localized, minute components, such as specifically shaped headlights or grilles. Under a shallow network architecture, the low-frequency structural enhancement branch tends to focus more on capturing global rigid contours. However, once transitioned to a deeper network, FEDSNet's adaptive weighting mechanism effectively compensates for this limitation and achieves a reversal in performance, further validating the universality and flexibility of the framework across different semantic abstraction levels.

\subsection{Ablation Studies}

To investigate the contributions of the core components in FEDSNet to fine-grained feature learning, an incremental ablation study is conducted on the Stanford Cars dataset. By constructing representative model variants, this study systematically validates the efficacy of each algorithmic module. The specific experimental results are presented in Table~\ref{tab2}.

\begin{table}[H]
\caption{Ablation study comparison of FEDSNet modules on the Stanford Cars dataset utilizing different backbone networks.\label{tab2}}
\centering
\begin{tabularx}{\textwidth}{L C C C}
\toprule
\textbf{Variant} & \textbf{Configuration} & \textbf{Conv-4 (1-shot)} & \textbf{ResNet-12 (5-shot)} \\
\midrule
V0 & DSN(Baseline) & $40.17\%$ & $94.03\%$ \\
V1 (+Freq) & Frequency Branch (1:1 Mean Fusion) & $52.13\%$ & $95.32\%$ \\
V2 (+Attn) & Frequency Attention (1:1 Mean Fusion) & $49.36\%~(\downarrow)$ & $94.27\%~(\downarrow)$ \\
V3 (FEDSNet) & Adaptive Gating Fusion & $\mathbf{53.19\%}$ & $\mathbf{95.75\%}$ \\
\bottomrule
\end{tabularx}
\end{table}

Variant V0, serving as the baseline DSN model, constructs a single subspace solely utilizing the raw spatial features. Its 1-shot accuracy under the Conv-4 backbone is $40.17\%$, reflecting that fine-grained targets in shallow features are highly susceptible to background noise interference. Building upon this, Variant V1 introduces the frequency branch (+Freq) and adopts a fixed 1:1 mean fusion strategy. Its performance increases to $52.13\%$ under Conv-4 and reaches $95.32\%$ under ResNet-12. This indicates that explicitly stripping high-frequency noise and extracting stable low-frequency structural features via the DCT transformation and low-pass filtering plays a crucial role in calibrating the principal directions of the subspace. However, when the frequency channel attention mechanism (+Freq-Attn) is introduced in Variant V2 while maintaining the fixed-ratio fusion, a noticeable performance drop is observed, with the Conv-4 accuracy falling to $49.36\%$ and ResNet-12 dropping to $94.27\%$.

This phenomenon does not indicate a flaw in the attention mechanism itself, but rather highlights the dynamic sensitivity of feature weights. The enhanced frequency features create an asymmetrical distribution scale relative to the spatial features. A forced mean fusion causes the high-energy frequency components to inadvertently overshadow the critical texture details within the spatial branch. Variant $V_3$, representing the complete FEDSNet model, introduces the adaptive gating fusion module to dynamically learn the weight coefficients $\alpha$ of the two views, enabling the Conv-4 performance to rebound and reach $53.19\%$. This performance recovery from $V_2$ to $V_3$ clearly demonstrates the necessity of deeply coupling frequency enhancement with adaptive weight adjustment. It confirms that the adaptive mechanism can dynamically find an optimal balance between capturing local textures and extracting global contours based on task requirements. Furthermore, the discriminative regularization term $L_{\mathrm{disc}}$ introduced in the optimization objective further enlarges the inter-class distance by forcing the basis vectors of different categories to be maximally orthogonal, ensuring that the generated dual-view subspaces possess stronger discriminability and generalization robustness.

\subsection{Complexity and Efficiency Analysis}

To evaluate the engineering practicality and operational efficiency of the proposed Frequency-Enhanced Dual-Subspace Network (FEDSNet), this section conducts a complexity comparison between the baseline DSN and the FEDSNet model under unified hardware environments and task settings. The test hardware is a single NVIDIA RTX 3090 GPU, the input resolution is set to $84 \times 84$, and the evaluation task adopts the standard $5$-way $5$-shot setting (a single task totals 100 input images). To verify the scalability of the algorithm across different depth architectures, we evaluate the model parameters (Params), theoretical floating-point operations (FLOPs), intermediate memory overhead (Forward/Backward Pass Size), peak GPU memory (Peak Mem), and average inference time per task (Time/Task) on both Conv-4 and ResNet-12 backbone networks. The detailed results are shown in Table~\ref{tab3}.

\begin{table}[H]
\caption{Complexity and efficiency comparison between DSN and FEDSNet under different backbone networks.\label{tab3}}
\centering
\footnotesize 
\setlength{\tabcolsep}{3pt} 
\begin{tabularx}{\textwidth}{l l C C C C C}
\toprule
\textbf{Backbone} & \textbf{Model} & \textbf{\makecell{Params\\(M)}} & \textbf{\makecell{FLOPs\\(G)}} & \textbf{\makecell{PassSize\\(MB)}} & \textbf{\makecell{PeakMem\\(MB)}} & \textbf{\makecell{Time/Task\\(ms)}} \\
\midrule
\multirow{2}{*}{Conv-4} & DSN (Baseline) & $0.11$ & $9.96$ & $914.16$ & $361.17$ & $114.54$ \\
 & FEDSNet (Ours) & $5.24$ & $10.34$ & $915.18$ & $381.20$ & $277.15$ \\
\midrule
\multirow{2}{*}{ResNet-12} & DSN (Baseline) & $12.42$ & $352.30$ & $5730.86$ & $694.05$ & $125.46$ \\
 & FEDSNet (Ours) & $13.30$ & $352.36$ & $5732.23$ & $697.39$ & $147.28$ \\
\bottomrule
\end{tabularx}
\end{table}

From the comparison in Table~\ref{tab3}, it is evident that FEDSNet maintains a reasonable spatial complexity and memory footprint; its core modules---frequency decoupling and dual-view reconstruction---introduce minimal computational overhead. For instance, under ResNet-12, the intermediate memory and peak memory only slightly increase by $1.37$ MB and $3.34$ MB, respectively.

Moreover, model efficiency is closely related to feature dimensionality. Under the Conv-4 setting, the high-dimensional $1600$-D features increase the SVD computation and CPU-GPU heterogeneous communication overhead, causing the inference time to rise to $277.15$~ms. However, in the ResNet-12 architecture, the compact $640$-D features generated by Global Average Pooling (GAP) effectively mitigate this issue. The total parameter count increases by less than $1$~M, and while keeping the FLOPs almost unchanged, the single-task inference time is controlled at $147.28$~ms, an increase of merely $21.8$~ms compared to the baseline.

In summary, FEDSNet trades a minor efficiency compromise for a robust structural discrimination mechanism, achieving a favorable balance between lightweight design and classification performance. Furthermore, it naturally adapts to modern deep feature extraction architectures, demonstrating practical potential for real-world deployment.

\section{Conclusion}

To address the persistent challenges of feature entanglement, susceptibility to background noise interference, and the structural instability of single-subspace metrics in few-shot fine-grained image classification, this paper proposes the Frequency-Enhanced Dual-Subspace Network (FEDSNet). Departing from traditional metric learning paradigms that rely exclusively on spatial-domain features, this framework introduces the Discrete Cosine Transform (DCT) and a low-pass filtering mechanism to explicitly decouple robust low-frequency global structural components from high-frequency environmental clutter. Subsequently, independent linear subspaces with low-rank constraints are constructed for both the spatial texture features and the frequency structural features. By designing an adaptive gating mechanism to dynamically fuse the dual-view projection distances, this deep coupling of frequency calibration and spatial metrics enables the model to dynamically achieve an optimal balance between capturing local textures and extracting global morphology.

Extensive experiments conducted on four mainstream fine-grained benchmark datasets—CUB-200-2011, Stanford Cars, Stanford Dogs, and FGVC-Aircraft—validate the effectiveness of the proposed method. The results demonstrate that under both shallow networks with limited representation capacity (Conv-4) and deep architectures (ResNet-12), FEDSNet exhibits highly competitive classification accuracy and strong robustness, consistently outperforming the baseline Deep Subspace Network (DSN) and various state-of-the-art few-shot learning models. Detailed ablation studies further confirm that FEDSNet achieves a favorable balance between performance and computational overhead. By utilizing a concise and efficient low-rank manifold architecture, the proposed framework effectively mitigates the overfitting phenomenon prevalent under extreme few-shot conditions.

While the proposed dual-view subspace framework has demonstrated significant effectiveness in suppressing background noise, it exhibits minor limitations when applied to fine-grained recognition tasks that heavily rely on minute, localized components (e.g., specific headlights or bird beaks). In such cases, a uniform low-pass filtering strategy may inadvertently discard subtle discriminative structural details. Consequently, future work will explore more refined, multi-band adaptive frequency partitioning mechanisms and consider integrating this frequency decoupling framework with local attention alignment techniques. Furthermore, extending this low-rank dual-subspace metric paradigm to broader downstream computer vision tasks, such as few-shot fine-grained image retrieval and video action recognition, remains a promising avenue for future research.

\authorcontributions{Conceptualization, G.W., M.W. and W.Z.; methodology, G.W. and M.W.; software, G.W.; validation, G.W., H.C. and B.Y.; formal analysis, G.W.; investigation, G.W.; resources, M.W. and W.Z.; data curation, G.W.; writing---original draft preparation, G.W.; writing---review and editing, W.Z., M.W., Y.W. and J.Y.; visualization, G.W.; supervision, M.W. and W.Z.; project administration, W.Z.; funding acquisition, M.W. All authors have read and agreed to the published version of the manuscript.}

\funding{This research was funded by the Project Supported by Scientific Research Program Funded by Shaanxi Provincial Education Department (Program No. 22JK0303) and Natural Science Basic Research Plan in Shaanxi Province of China (Program No. 2022JQ-175). The APC was funded by the same grants.}

\institutionalreview{Not applicable.}

\informedconsent{Not applicable.}

\dataavailability{The datasets used in this study are publicly available fine-grained benchmark datasets. CUB-200-2011 is available at \url{https://www.vision.caltech.edu/datasets/cub_200_2011/}, Stanford Cars at \url{http://ai.stanford.edu/~jkrause/cars/car_dataset.html}, Stanford Dogs at \url{http://vision.stanford.edu/aditya86/ImageNetDogs/}, and FGVC-Aircraft at \url{https://www.robots.ox.ac.uk/~vgg/data/fgvc-aircraft/}. The source code supporting the reported results is available from the corresponding author upon reasonable request.}

\acknowledgments{The authors thank the School of Electronic Information and Artificial Intelligence, Shaanxi University of Science and Technology for providing computational resources for this research.}

\conflictsofinterest{The authors declare no conflicts of interest.}
\reftitle{References}

\end{document}